\documentclass{article}

\usepackage{PRIMEarxiv}

\usepackage[utf8]{inputenc} 
\usepackage[T1]{fontenc}    
\usepackage{hyperref}       
\usepackage{url}            
\usepackage{booktabs}       
\usepackage{amsfonts}       
\usepackage{nicefrac}       
\usepackage{microtype}      
\usepackage{lipsum}
\usepackage{fancyhdr}       
\usepackage{graphicx}       
\graphicspath{{media/}}     
\usepackage{hyperref}
\usepackage{rotating} 
\usepackage{fontawesome5}
\hypersetup{hidelinks=true}
\usepackage{tikz}
\usepackage{epigraph}
\usepackage[shortlabels]{enumitem}
\usepackage{xcolor,colortbl}
\usepackage{array, makecell, multirow, tabularx}
\usepackage{multicol}
\usepackage{threeparttable}
\usepackage{rotating}
\usepackage{nicematrix}
\usepackage{adjustbox}
\usepackage{framed}
\usepackage{color,soul}
\usepackage[most]{tcolorbox}
\usepackage{subcaption} 
\usepackage{listings}
\usepackage{academicons}
\usepackage{enumitem}
\usepackage{float}
\usepackage{orcidlink}
\usepackage{array}  

\usepackage{booktabs}
\usepackage{pifont}
\lstdefinestyle{myStyle}{
    belowcaptionskip=1\baselineskip,
    breaklines=true,
    frame=none,
    numbers=none,
    basicstyle=\footnotesize\ttfamily,
    keywordstyle=\bfseries\color{green!40!black},
    commentstyle=\itshape\color{purple!40!black},
    identifierstyle=\color{blue},
    backgroundcolor=\color{gray!10!white},
}

\definecolor{codegreen}{rgb}{0,0.6,0}
\definecolor{codegray}{rgb}{0.5,0.5,0.5}
\definecolor{codepurple}{rgb}{0.5,0,0.82}
\definecolor{backcolour}{rgb}{0.95,0.95,0.92}
\definecolor{darkblue}{rgb}{0.0, 0.0, 0.55}
\definecolor{bluesky}{rgb}{0.53, 0.81, 0.98}
\definecolor{codegreen}{rgb}{0,0.6,0}
\definecolor{backcolour}{rgb}{0.95,0.95,0.92}

\usepackage{capt-of}
\usepackage{verbatim}
\usepackage{outlines}
\usepackage{cleveref}
\usepackage[numbers]{natbib}
\usepackage{rotating}

\usepackage[group-separator={,}]{siunitx}
\title{ \raisebox{-0.3\height}{\includegraphics[width=0.06\linewidth]{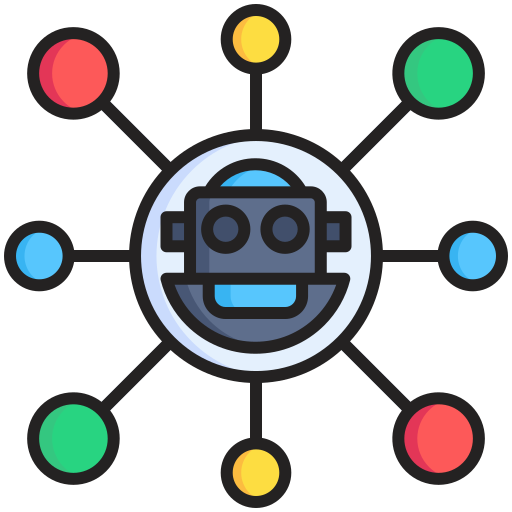}} \textbf{Combinatorial Optimization for All:} \\Using LLMs to Aid Non-Experts in Improving Optimization Algorithms}

\author{
  Camilo Chacón Sartori \orcidlink{0000-0002-8543-9893}\thanks{Corresponding author: \texttt{cchacon@iiia.csic.es} }\\
  Artificial Intelligence Research Institute (IIIA-CSIC)\\
  Bellaterra, Spain\\
  \texttt{cchacon@iiia.csic.es}
  \and
  Christian Blum \orcidlink{0000-0002-1736-3559}\\
  Artificial Intelligence Research Institute (IIIA-CSIC)\\
  Bellaterra, Spain\\
  \texttt{christian.blum@iiia.csic.es}
}

\begin{document}
\maketitle

\vspace{-3em}
\begin{center}
  \raisebox{-0.3\height}{\includegraphics[width=0.03\textwidth]{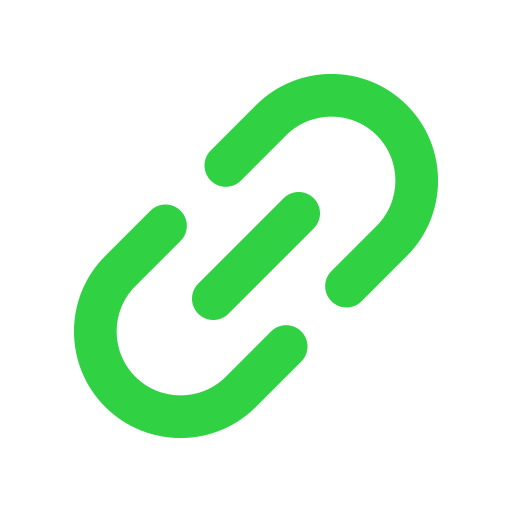}}  Project URL: \textcolor{darkblue}{\url{https://camilochs.github.io/comb-opt-for-all/}}
\end{center}

\begin{figure}[h]
\centering
  \includegraphics[width=0.9\linewidth]{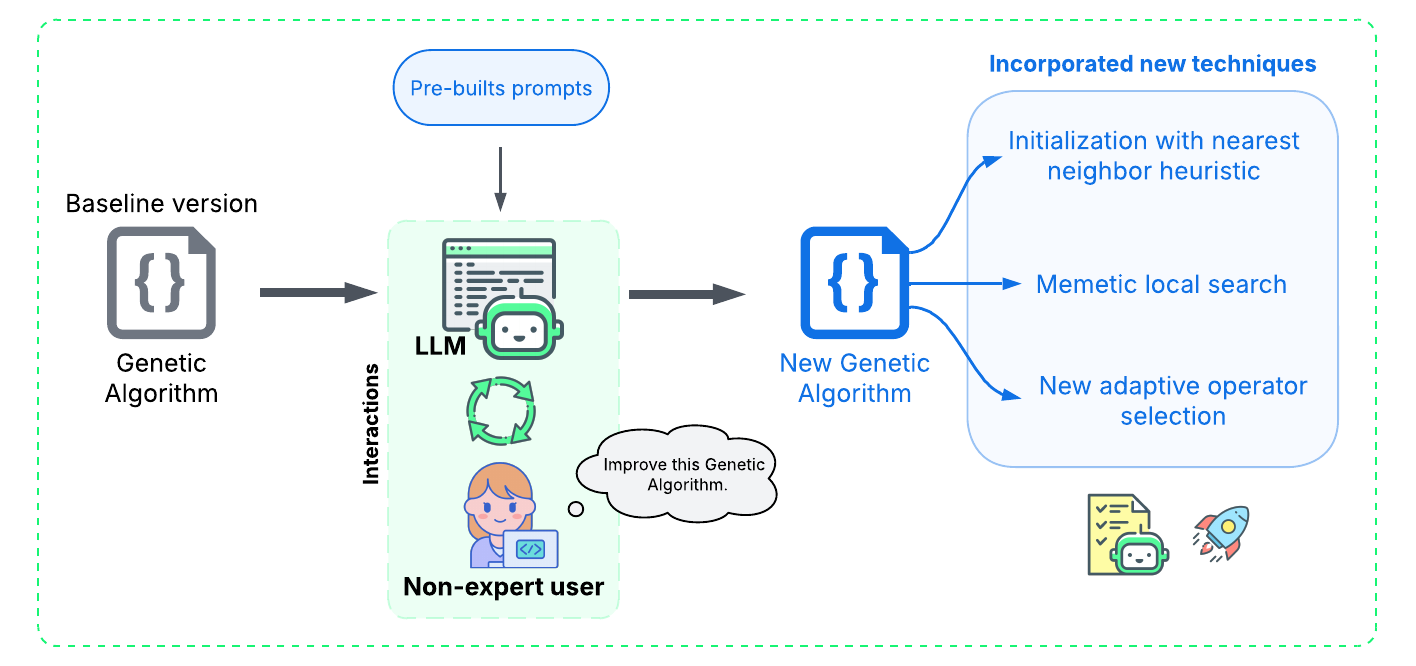}
  \caption{A non-expert user's interaction with an LLM can enhance an existing genetic algorithm by incorporating modern techniques.}
  \label{fig:example}
\end{figure}

\begin{abstract}
Large Language Models (LLMs) have shown notable potential in code generation for optimization algorithms, unlocking exciting new opportunities. This paper examines how LLMs, rather than creating algorithms from scratch, can improve existing ones without the need for specialized expertise. To explore this potential, we selected 10 baseline optimization algorithms from various domains (metaheuristics, reinforcement learning, deterministic, and exact methods) to solve the classic Travelling Salesman Problem. The results show that our simple methodology often results in LLM-generated algorithm variants that improve over the baseline algorithms in terms of solution quality, reduction in computational time, and simplification of code complexity, all without requiring specialized optimization knowledge or advanced algorithmic implementation skills.
\end{abstract}


\section{Introduction}
If we asked you how many optimization algorithms exist, would you be able to come up with an exact answer? Probably not, as there are simply too many algorithms (or algorithm variants) to count. A simple search for `optimization algorithm' in databases like Scopus, IEEE Xplore, or GitHub returns thousands of results. And that is just the start. Classic algorithms, such as the Genetic Algorithm, have spawned so many variations that some barely resemble the original. On top of that, hybrid approaches combine heuristics with exact methods or machine learning techniques. New algorithms are introduced every day.

All these algorithms---whether open-source or proprietary---can be improved using modern technologies. Enhancing their implementations with advanced techniques would benefit not only optimization specialists but also non-expert users. In this paper, we define ``non-expert'' users as those who have programming skills but lack formal training in optimization theory. 

The emergence of Large Language Models (LLMs)---AI systems trained on vast text corpora that can generate code and assist with complex tasks---showcased by innovations like OpenAI's GPT-O1~\cite{openai2024gpt4technicalreport}, Anthropic's Claude~\cite{anthropicIntroducingClaude}, Google's Gemini~\cite{geminiteam2024gemini15unlockingmultimodal}, Meta's Llama~\cite{grattafiori2024llama3herdmodels}, and DeepSeek R1~\cite{deepseekai2025deepseekr1incentivizingreasoningcapability}, has significantly broadened possibilities for scientific innovation. These models demonstrate remarkable code generation capabilities through tools like GitHub Copilot\footnote{\url{https://github.com/features/copilot}} and Cursor AI\footnote{\url{https://www.cursor.com/}}, enhancing developer efficiency. This raises an intriguing possibility: could individuals without optimization expertise leverage LLMs to enhance existing optimization algorithms? 

Current research on LLMs for algorithm design has largely followed two paths: evolving novel heuristics or specific components within complex frameworks~\cite{Romera-Paredes2024, 10752628, ye2024reevolargelanguagemodels, liu2024evolutionheuristicsefficientautomatic, dat2024hsevoelevatingautomaticheuristic}, and discovering specific heuristic operators by leveraging the LLM’s semantic understanding of the code to identify parameters that are unused in one part but could be repurposed or utilized in another~\cite{sartori2025improvingexistingoptimizationalgorithms}. However, less attention has been given to the practical challenge of improving \textit{complete existing algorithm implementations as a whole}, without requiring theoretical knowledge of combinatorial optimization.

To bridge this gap, our work presents the first large-scale, systematic evaluation of the ability of LLMs to upgrade classical algorithms across a wide range of families---including metaheuristics, reinforcement learning, deterministic heuristics, and exact methods. Our study focuses on the canonical Travelling Salesman Problem (TSP), where we apply a simple and reproducible prompting strategy through a chatbot interface. While our experiments are centered on the TSP, its foundational nature as a problem of optimal sequencing makes it a highly relevant case study. The core challenge of finding the best permutation of nodes is shared by many other classic optimization problems, such as the Vehicle Routing Problem, the Hamiltonian Cycle Problem, and the Sequential Ordering Problem. Therefore, the strategies and findings from this work are likely applicable to any problem centered on the fundamental question: ``In what order should a series of tasks be performed to minimize total cost?''

Our findings demonstrate that LLMs can propose and implement sophisticated improvements, such as incorporating modern heuristic components or engineering more efficient data structures, often resulting in a reduction in code complexity compared to the original implementations. Crucially, these enhancements are applied not to isolated code fragments (heuristics), as in previous work, but to the entirety of an optimization algorithm's codebase. Furthermore, we analyze cases where the improvements were not substantial, offering insights into the current limitations of this approach. Ultimately, this work showcases a practical methodology that lowers the barrier for practitioners without deep optimization expertise to access higher-performance algorithms, leveraging the vast knowledge embedded in modern LLMs. A schematic of our framework using a Genetic Algorithm is shown in Figure~\ref{fig:example}.

The paper unfolds as follows. In Section~\ref{sec:background}, we provide an overview of LLM advancements in combinatorial optimization, introduce the TSP problem, and briefly describe the 10 selected algorithms. Our methodology for enhancing existing optimization algorithms is detailed in Section~\ref{sec:methodology}. An exhaustive analysis of the 10 optimization algorithms improved through LLM application is presented in Section~\ref{sec:evaluation}. Section~\ref{sec:disc} discusses implications and identifies directions for future research. We conclude by highlighting the key findings from our investigation.

\begin{sidewaystable}[htbp]
\centering
\caption{Expanded Comparative Analysis of LLM-based Algorithm Design Approaches}
\label{tab:expanded_comparative_analysis}
\begin{tabularx}{\textheight}{@{} l X | X X X X @{}}
\toprule
\textbf{Feature} & \textbf{Our Work} & \textbf{Improving Existing Optimization Algorithms~\cite{sartori2025improvingexistingoptimizationalgorithms}} & \textbf{AlphaEvolve~\cite{Novikov2025-vg}} & \textbf{LLaMEA/ReEvo~\cite{10752628, ye2024reevolargelanguagemodels}} & \textbf{FunSearch/Evo-Heuristics~\cite{Romera-Paredes2024, liu2024evolutionheuristicsefficientautomatic}} \\
\midrule
\textbf{Main Goal} & 
To provide a \textbf{large-scale benchmark} on enhancing \textit{complete} implementations of \textit{existing} algorithms from diverse families. &
To demonstrate a \textbf{proof of concept} through the improvement of a \textit{single}, complex heuristic in a state-of-the-art algorithm. &
To autonomously \textbf{evolve a population} of algorithms starting from a seed program, leveraging LLM-based agents to discover improved variants.&
To \textbf{generate new} metaheuristics or heuristic components using an evolutionary computation loop. &
To \textbf{discover specific} mathematical functions or heuristics from scratch. \\
\addlinespace
\textbf{LLM Input} & 
A complete, functional code file from a well-known framework~\cite{pereira2022pycombinatorial}. &
The complete code of a single, high-performance algorithm. &
An initial ``seed'' algorithm and a fitness function. &
A set of prompts representing algorithm components (e.g., operators). &
A problem description and a code skeleton for a specific function. \\
\addlinespace
\textbf{Improvement Process} & 
Interactive LLM-driven prompting and refinement of full algorithms. &
Interactive LLM prompting to refine a specific heuristic or function. &
Evolutionary loop where the LLM acts as an advanced mutation operator to generate new programs. &
Evolutionary loop where the LLM acts as a crossover/mutation operator on algorithm components. &
Program search tree where the LLM proposes new function implementations. \\
\addlinespace
\textbf{User's Role} & 
\textbf{Practitioner/Non-expert}: Provides base code, validates final solution. &
\textbf{Expert}: Seeks to enhance a specific, already-strong algorithm. &
\textbf{Expert}: Designs the evolutionary process and fitness function. &
\textbf{Expert}: Designs the evolutionary framework and component prompts. &
\textbf{Expert}: Defines the problem, evaluator, and code structure. \\
\addlinespace
\textbf{Key Contribution} & 
A \textbf{systematic benchmark} demonstrating that a single, prompt-based methodology can holistically improve algorithms from diverse families without specialized theoretical knowledge. &
A \textbf{case study} demonstrating the feasibility of LLM-based enhancement on a complex, expert-level algorithm. &
An \textbf{evolutionary framework} for automated program search and algorithm improvement. &
A \textbf{framework} for the automatic generation of algorithms. &
A \textbf{method} for the automatic discovery of functions/heuristics. \\
\bottomrule
\end{tabularx}
\end{sidewaystable}

\section{Background}\label{sec:background}
\subsection{Large Language Models in Combinatorial Optimization}\label{subsec:background}
Large Language Models (LLMs) have recently shown promise in optimization tasks~\cite{yang2024largelanguagemodelsoptimizers, liu2024largelanguagemodelsevolutionary, huang2024largelanguagemodelmeets} by exploiting the vast knowledge gained during their pre-training phase. Beyond guiding the optimization process, LLMs excel at detecting patterns, identifying key features in problem instances, and refining search spaces. They have also shown to be able to generate new heuristics tailored to specific problems. Furthermore, LLMs offer valuable insights by explaining the results of optimization problems, making them versatile tools for both solving and interpreting complex tasks.

In this article, we focus on a specific type of optimization problem: combinatorial problems. These problems have unique properties that set them apart: many valid solutions (including the existence of equivalent or similar solutions), decomposability (some problems can be broken down into smaller, more manageable subproblems), constraint handling (a set of rules that define valid solutions), and search space structure (the presence of multiple local optima, requiring well-chosen search strategies to avoid getting trapped in suboptimal solutions), among others.  

As a result, researchers in this field must not only be knowledgeable about combinatorial optimization problems but also highly proficient in implementing optimization algorithms. Given the importance of computational efficiency, factors such as memory optimization, effective data structure management, minimizing unnecessary abstractions, and carefully selecting the right programming language play a crucial role in the design and development of these algorithms.

\begin{table*}[tb]
    \centering
    \caption{Overview of Selected Algorithms for Solving the Travelling Salesman Problem (TSP)}
    \resizebox{\linewidth}{!}{
    
\renewcommand{\arraystretch}{1.5}
    \begin{tabular}{l|l|l}
        \toprule
        \textbf{Algorithm} & \textbf{Characteristics} & \textbf{Application to TSP} \\
        \midrule
        \multicolumn{3}{l}{ \textcolor[RGB]{70, 130, 180}{\textbf{Metaheuristic}}} \\
        Ant Colony Optimization (ACO)~\cite{585892} & Probabilistic, pheromone-based learning & Simulates ants' foraging behavior where solutions (routes) are constructed based on pheromone trails left by previous solutions. \\
        Genetic Algorithm (GA)~\cite{Potvin1996} & Population-based, crossover, mutation & Generates a population of routes and evolves them through selection, crossover, and mutation to find near-optimal solutions. \\
        Adaptive Large Neighborhood Search (ALNS)~\cite{Ropke2006} & Adaptive destruction and reconstruction & Iteratively destroys and reconstructs routes, dynamically adjusting strategies based on previous performance. \\
        Tabu Search (TABU)~\cite{Glover1989TabuS} & Use of memory structures (tabu lists) & Iteratively modifies routes while keeping a list of features of previously visited solutions to prevent revisits. \\
        Simulated Annealing (SA)~\cite{Kirkpatrick1983} & Probabilistic, temperature-based search & Iteratively refines a route by accepting worse solutions with a decreasing probability to escape local optima. \\
        
        \midrule
        \multicolumn{3}{l}{\textcolor[RGB]{70, 130, 180}{\textbf{Reinforcement Learning}}} \\
        Q\_Learning~\cite{Wang2023ReinforcementLF} & Value-based learning, exploration-exploitation trade-off & Learns an optimal routing policy by iteratively updating action-value functions based on rewards from different paths. \\
        SARSA~\cite{Wang2023ReinforcementLF} & On-policy learning, continuous updates & Uses an on-policy approach to learning optimal routing strategies based on real-time interactions with the environment. \\

        \midrule
        \multicolumn{3}{l}{\textcolor[RGB]{70, 130, 180}{\textbf{Deterministic Heuristic}}} \\
        Christofides~\cite{Christofides2022} & Guarantees 1.5-optimality, MST-based & Constructs a minimum spanning tree, finds perfect matching, and combines them to form a tour. \\
        Convex Hull~\cite{4309370} & Geometric approach & Starts with a convex hull and incrementally inserts remaining points in a way that minimizes travel distance. \\
        
        \midrule
        \multicolumn{3}{l}{\textcolor[RGB]{70, 130, 180}{\textbf{Exact}} } \\
        Branch and Bound (BB)~\cite{MORRISON201679} & Systematic enumeration, pruning & Explores all possible solutions while pruning suboptimal paths to guarantee optimality. \\
        \bottomrule
    \end{tabular}}
    \label{tab:algorithms}
\end{table*}
  
The rapid advancements in using LLMs as black-box collaborators for optimization demand a clear positioning of new methodologies. Table~\ref{tab:expanded_comparative_analysis} delineates our work's unique contributions in this crowded landscape. Specifically, our approach is distinguished by:

\begin{itemize}
    \item \textbf{Systematic Scope over Anecdotal Evidence.} Where prior work often provides proofs-of-concept on a limited set of algorithms, we deliver a systematic benchmark across ten algorithms spanning four distinct families. We prove that a straightforward, reproducible prompting strategy is robust enough to yield significant performance gains across this diverse set without compromising code stability.

    \item \textbf{Practitioner-Centric Design over Expert-Centric Frameworks.} Our methodology marks a fundamental departure from expert-centric systems that require designing complex evolutionary loops or prompt engineering strategies. It is uniquely designed for the practitioner---a user with programming skills but lacking deep theoretical expertise. Our work is the first to demonstrate a path for these users to upgrade complete, monolithic codebases, rather than just optimizing isolated functions. This holistic approach significantly lowers the barrier for applying state-of-the-art optimization techniques.

\end{itemize}

To introduce this novel research direction, the next two subsections present a classic combinatorial optimization problem along with ten traditional optimization algorithms, grouped into four distinct categories. Then, in the methodology section, Section~\ref{sec:methodology}, we demonstrate how LLMs can be leveraged to enhance these ten algorithms, improving both performance and efficiency while streamlining their implementations.

\subsection{The Travelling Salesman Problem: A Brief Description}\label{subsec:problem}

The Travelling Salesman Problem (TSP) is one of the most iconic problems in combinatorial optimization, serving as a foundational pillar in both Artificial Intelligence and Operations Research. Formally, let $\mathcal{C} = \{c_1, c_2, \dots, c_n\}$ be a set of \( n \) cities, and let $D: \mathcal{C} \times \mathcal{C} \to \mathbb{R}_{\geq 0}$ be a function that assigns a non-negative distance between each pair of cities. The objective of the TSP is to find a permutation \( \pi \) of the indices \( \{1,2,\dots,n\} \) that minimizes the total travel distance of a closed tour, formally defined as:
\[
\min_{\pi \in S_n} \left\{ \sum_{i=1}^{n-1} D(c_{\pi(i)}, c_{\pi(i+1)}) + D(c_{\pi(n)}, c_{\pi(1)}) \right\},
\]
where \( S_n \) denotes the set of all permutations of \( \{1,2,\dots,n\} \).

For this study, we selected the TSP because of the vast amount of implementations available online---public code repositories, books, and scientific articles---indicating that LLMs likely possess extensive knowledge of techniques for solving it~\cite{ma2023trainingstagedoescode}.

\subsection{Traditional Optimization Algorithms for the TSP}\label{subsec:algorithms}

Algorithms for solving the TSP come in a broad variety of forms and approaches. We have chosen ten distinct optimization algorithms, summarized in Table~\ref{tab:algorithms}. These algorithms are categorized into four groups: metaheuristic methods (stochastic, iterative optimization algorithms), reinforcement learning (policy-driven), deterministic heuristics, and an exact algorithm. These four algorithm categories can be briefly characterized as follows:
\begin{enumerate}
    \item \textbf{Metaheuristics}~\cite{blum2003metaheuristics} are versatile optimization methods that use heuristic and stochastic principles to search solution spaces. While not guaranteeing global optima, they efficiently find high-quality solutions. 
    \item \textbf{Reinforcement Learning}~\cite{10.5555/2670001} is a machine learning paradigm where an agent learns optimal decision-making by interacting with an environment and maximizing cumulative rewards. 
    \item \textbf{Deterministic heuristics}~\cite{martí2011linear} are among the most basic algorithms for combinatorial optimization. They generate generally one solution from scratch by making a myopic, deterministic decision at each step. 
    \item \textbf{Exact algorithms}~\cite{martí2011linear} ensure optimality by exhaustively exploring the solution space. 
\end{enumerate}
The choice of algorithms from these four categories guarantees that our LLM-based improvement framework is tested across a diverse spectrum of methodologies, as the selected algorithms---though addressing the same problem (TSP)---vary fundamentally in their underlying principles.

\subsection{Selected Implementations}
To minimize the possibility of implementation errors in the ten algorithms for solving the TSP, and to ensure that these implementations are being utilized by the community, we employed \texttt{pyCombinatorial}~\cite{pereira2022pycombinatorial}.\footnote{\url{https://github.com/Valdecy/pyCombinatorial}} This Python library, which includes a whole range of optimization algorithms for solving the TSP, has received over 100 stars on GitHub and was created by \href{https://scholar.google.com/citations?user=YPvpg9UAAAAJ&hl=pt-BR}{Valdecy Pereira}. Each implementation is based on a standard algorithm variant, providing us with an excellent testing environment for attempting to improve them using LLMs. In the next section, we detail our methodology for improving optimization algorithms.

\section{Methodology}\label{sec:methodology}
\subsection{Enhancing Traditional Optimization Algorithms with Large Language Models}

We introduce a methodology based on LLM interactions to generate enhanced versions of the 10 before-mentioned algorithms, building upon and extending the approach proposed by~\cite{sartori2025improvingexistingoptimizationalgorithms} (see Figure~\ref{fig:example}). This process can be replicated using the chatbot available at the project URL.\footnote{\url{https://camilochs.github.io/comb-opt-for-all/}}

Given the initial set $\{A_i \mid i=1,\ldots,10\}$ of 10 original algorithm codes and an LLM, the algorithm improvement process can technically be stated as follows. First, a prompt $P_i$ is generated based on our general prompt template $T$ (shown in the second column of this page), the algorithm name $N_i$, the signature of the main function $S_i$, and the algorithm code $A_i$:
\begin{equation}\label{eq:prompt_production}
P_i = \textsc{Produce\_Prompt}(T, N_i, S_i, A_i), \quad i = 1, \dots, 10
\end{equation}
Then, the generated prompt $P_i$ is executed by the LLM given a set of hyperparameters $\theta$, resulting in a changed/updated implementation $A'_i$:
\begin{equation*}
A'_i = \text{LLM}_{\text{execute}}(P_i, \theta), \quad i = 1, \dots, 10
\end{equation*}

To ensure correctness, each \( A'_i \) undergoes a validation process (see below). If an output fails validation, the refinement process is repeated iteratively until a valid version is obtained.

\subsubsection{Code Validation}  

The validation of \( A'_i \) may fail for two reasons:  
\begin{enumerate}  
    \item \textbf{Execution errors}: These lead to immediate failures during code runtime.  
    \item \textbf{Logical inconsistencies}: The algorithm executes without errors but produces invalid TSP solutions.  
\end{enumerate}  

In the first case, an error message \( e \) is generated, and the LLM refines the code based on this feedback:  
\[
A'_i = \text{LLM}_{\text{execute}}(A'_i, \theta, e)
\]
For the second case, where execution is error-free but the generated solutions are invalid, an explicit prompt requesting a correction is passed to the LLM, such as:
\begin{center}
\textit{``The provided code generates invalid solutions; please verify and return a corrected version.''}  
\end{center}
The refinement loop proceeds until a valid code \( A'_i \) is obtained. This process is carried out through an interactive conversation with an LLM-based chatbot. To increase the chances of generating a valid code with each retry, we begin with a high-temperature setting, which is then progressively lowered in each iteration of the process. 

\begin{tcolorbox}[colback=bluesky!10,  boxrule=0.2pt, breakable]
In LLMs, \textit{temperature} controls the randomness of the model's output. A higher temperature (e.g., 1.0 or 2.0) makes the model's responses more diverse and less predictable, while a lower temperature (e.g., 0.2) makes the output more deterministic and stable.
\end{tcolorbox}

Table~\ref{tab:generation_algorithms} shows the results of this procedure for the five selected LLMs.\footnote{The reasoning behind selecting these five LLMs (rather than others) will be explained in Section~\ref{sec:evaluation}.} A green checkmark (\textcolor{green}{\faCheck}) indicates that the first obtained \( A'_i \) was valid, while a red cross (\textcolor{red}{\faTimes}) signifies that corrections were necessary due to either of the two code failures. A double red cross (\includegraphics[width=0.02\textwidth]{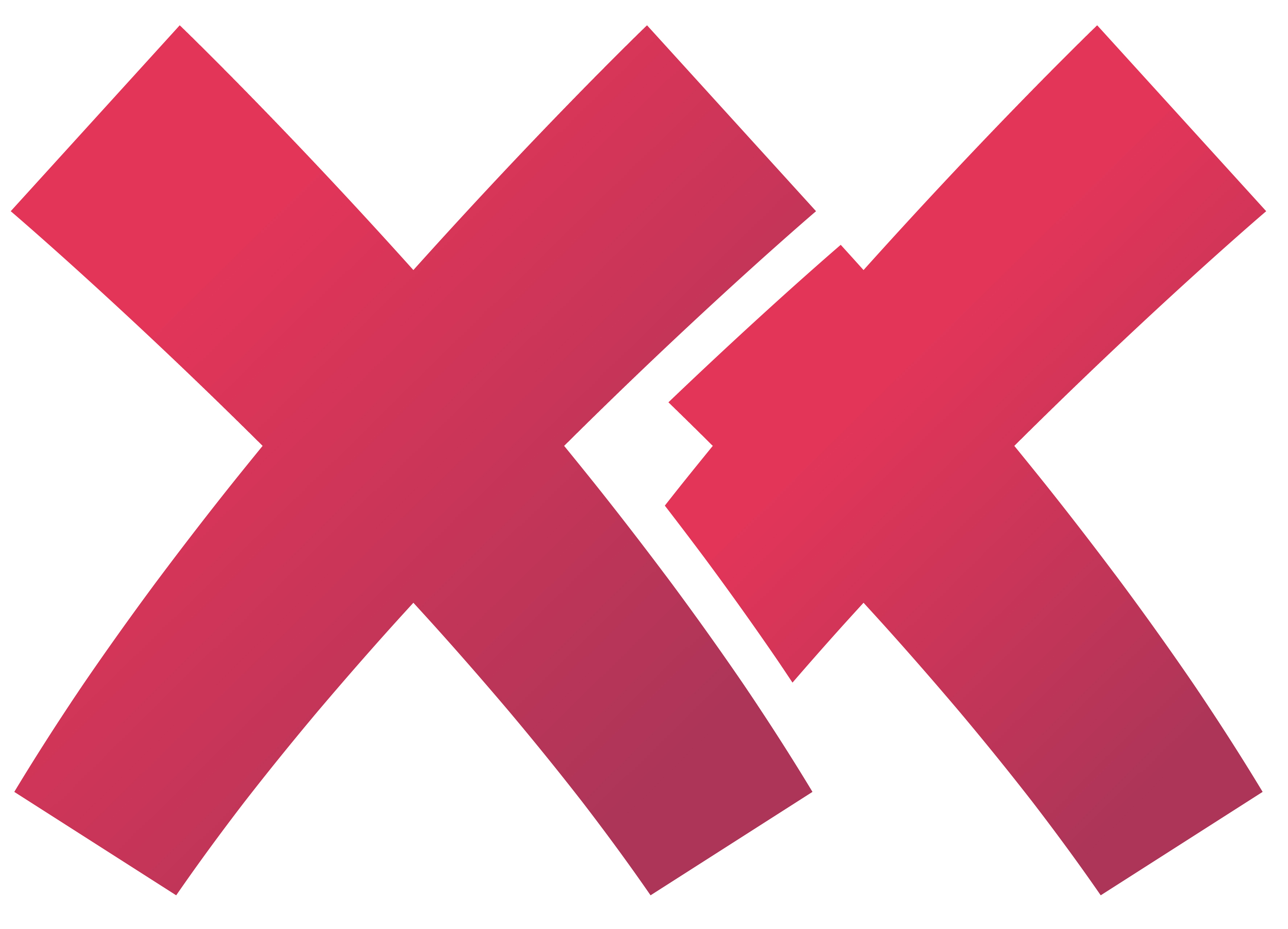}) indicates that both failures occurred. Moreover, in the case of corrections, the number of necessary corrections is provided in a second column. Note that Table~\ref{tab:generation_algorithms}, below the LLM names, also indicates the initial temperature setting.

\begin{table}[!t]
    \centering
    \caption{Analysis of the Code Generation Process}
    \resizebox{\columnwidth}{!}{ 
    \begin{tabular}{lcc|cc|cc|cc|cc}
        \toprule
        \multirow{4}{*}{\textbf{Algorithm}} & \multicolumn{2}{c}{\textbf{Claude-3.5-Sonnet}} & \multicolumn{2}{c}{\textbf{Gemini-exp-1206}} & \multicolumn{2}{c}{\textbf{Llama-3.3-70B}} & \multicolumn{2}{c}{\textbf{GPT-O1}} & \multicolumn{2}{c}{\textbf{DeepSeek-R1}} \\
        & \multicolumn{2}{c}{(temp = 1.0)} & \multicolumn{2}{c}{(temp = 2.0)} & \multicolumn{2}{c}{(temp = 1.0)} & \multicolumn{2}{c}{(temp = 1.0)} & \multicolumn{2}{c}{(temp = 1.0)} \\
        \cmidrule(lr){2-3} \cmidrule(lr){4-5} \cmidrule(lr){6-7} \cmidrule(lr){8-9} \cmidrule(lr){10-11}
        & \multicolumn{2}{c}{Success} & \multicolumn{2}{c}{Success} & \multicolumn{2}{c}{Success} & \multicolumn{2}{c}{Success} & \multicolumn{2}{c}{Success} \\
        & 1st Try & \# Attempts & 1st Try & \# Attempts & 1st Try & \# Attempts & 1st Try & \# Attempts & 1st Try & \# Attempts \\
        \midrule
        ACO   & {\color{red}\faTimes} & 1 & {\color{green}\faCheck} & - & {\color{green}\faCheck} & - & {\color{green}\faCheck} & - & {\color{green}\faCheck} & - \\
        GA    & {\color{green}\faCheck} & - & {\color{red}\faTimes} & 1 & {\color{green}\faCheck} & - & {\includegraphics[width=0.02\textwidth]{figures/error-doble.jpg} } & 3 & {\color{green}\faCheck} & - \\
        ALNS  & {\color{red}\faTimes} & 1 & {\color{green}\faCheck} & - & {\color{green}\faCheck} & - & {\color{green}\faCheck} & - & {\includegraphics[width=0.02\textwidth]{figures/error-doble.jpg} } & 3 \\
        TABU  & {\color{green}\faCheck} & - & {\color{green}\faCheck} & - & {\color{green}\faCheck} & - & {\color{green}\faCheck} & - & {\color{green}\faCheck} & - \\
        SA    & {\color{green}\faCheck} & - & {\color{green}\faCheck} & - & {\color{green}\faCheck} & - & {\color{green}\faCheck} & - & {\color{green}\faCheck} & - \\
        \midrule
        Q\_Learning & {\color{green}\faTimes} & - & {\color{green}\faCheck} & - & {\color{green}\faCheck} & - & {\color{green}\faCheck} & - & {\color{green}\faCheck} & - \\
        SARSA & {\color{green}\faTimes} & - & {\color{green}\faCheck} & - & {\color{green}\faCheck} & - & {\color{green}\faCheck} & - & {\color{green}\faCheck} & - \\
        \midrule
        Christofides   & {\color{green}\faTimes} & - & {\color{green}\faCheck} & - & {\color{green}\faCheck} & - & {\color{green}\faCheck} & - & {\color{green}\faCheck} & - \\
        Convex Hull   & {\color{red}\faTimes} & 1 & {\color{green}\faCheck} & - & {\color{green}\faCheck} & - & {\color{green}\faCheck} & - & {\color{green}\faCheck} & - \\
        \midrule
        Branch and Bound   & {\color{green}\faCheck} & - & {\includegraphics[width=0.02\textwidth]{figures/error-doble.jpg} } & 4 & {\color{green}\faCheck} & - & {\color{green}\faCheck} & - & {\color{green}\faCheck} & - \\
        \bottomrule
    \end{tabular}\label{tab:generation_algorithms}
    }
\end{table}

\subsection{Prompt Design: A Focus on Simplicity and Accessibility}

A central hypothesis of this study is that significant algorithmic improvements can be achieved without requiring users to possess deep expertise in prompt engineering. To test this, we developed a single, standardized prompt template. The design of this prompt is \textit {intentionally basic}.

Our objective is not to engage in an exhaustive search for the ``optimal'' algorithm variant through elaborate prompting, which constitutes a separate research direction. Instead, our goal is to establish a reproducible baseline for what LLMs can achieve when prompted by a non-expert user. This approach directly aligns with our focus on accessibility and allows us to isolate the LLM's intrinsic ability to enhance code.

Despite its simplicity, the prompt's formulation provides clear, high-level directives. It focuses the LLM on two critical performance axes---(1) improving solution quality and (2) accelerating convergence---and explicitly encourages the integration of state-of-the-art techniques to achieve these goals. The template is shown below:

\begin{tcolorbox}[
  colback=bluesky!10!white,        
  colframe=bluesky!70!black,       
  boxrule=0.3pt,                  
  arc=3mm,                      
  breakable,                    
  enhanced,                     
  title={\textbf{Prompt Template}},  
  fonttitle=\sffamily,    
  coltitle=white,               
  attach boxed title to top left={yshift=-0.1mm, xshift=2mm}, 
  boxed title style={colback=bluesky!70!black} 
]

\begin{lstlisting}[
  basicstyle=\ttfamily\small, 
  breaklines=true,
  frame=none,
  escapeinside={(*@}{@*)}
]
You are an optimization algorithm expert. 

I need to improve this (*@\textbf{\{\{algorithm name\}\}}@*) implementation for the travelling salesman problem (TSP) by incorporating state-of-the-art techniques. Focus on:

1. Finding better quality solutions
2. Faster convergence time

Requirements:
- Keep the main function signature: (*@\textbf{\{\{the signature of an the main function\}\}}@*)
- Include detailed docstrings explaining:
  * What improvement is implemented
  * How it enhances performance
  * Which state-of-the-art technique it is based on
- All explanations must be within docstrings, no additional text
- Check that there are no errors in the code

IMPORTANT: 
- Return ONLY Python code
- Any explanation or discussion must be inside docstrings
- At the end, include a comment block listing unmodified functions from the original code

Current implementation:
(*@\textbf{\{\{algorithm code\}\}}@*)
\end{lstlisting}

\end{tcolorbox}

The prompt template begins with \textit{``You are an optimization algorithm expert,''} a technique known as ``role prompting'', which has been empirically shown to guide LLMs toward a specific behavior or specialization~\cite{kong2024betterzeroshotreasoningroleplay, barkley2024investigatingrolepromptingexternal}. By setting a clear context from the outset, it enhances both the relevance and quality of the model’s response.  

This approach aligns with ``in-context learning''~\cite{dong2024surveyincontextlearning, Li2023-ij, schulhoff2025promptreportsystematicsurvey}, combining an external context (the user-provided code in the prompt) with an internal one (the LLM’s knowledge of various techniques to improve the TSP algorithm).

\begin{tcolorbox}[colback=bluesky!10,  boxrule=0.2pt, breakable]
\textbf{An important insight:} supplying a complete algorithm code within the prompt works like a `map,' guiding the model on how to update the code. The LLM must preserve the overall structure, making modifications only in the relevant sections without breaking the logic. Without this external context (the provided algorithm), the model’s solution would likely be more constrained and less effective, as it would have to generate everything from scratch. In contrast, with an initial codebase, the model can focus on refining and improving specific areas rather than rebuilding the entire algorithm code from scratch. In other words, the provided code \textit{influences} the update proposed by the LLM~\cite{dong2024surveyincontextlearning, sartori2025improvingexistingoptimizationalgorithms}.
\end{tcolorbox}

As technically indicated already in Eq.~\ref{eq:prompt_production}, the prompt template receives three dynamic variables that are placed in the corresponding positions in the prompt template (enclosed within \texttt{{\{\{ ... \}\}}}):  
\begin{itemize}
    \item \textbf{Algorithm's name}: steers the LLM toward a specific context.
    \item \textbf{Main function's signature}: ensures the initial function's input arguments, output values, and name remain unchanged, preventing unintended modifications that could affect compatibility with the original code.
    \item \textbf{Algorithm code}: the optimization algorithm's original implementation in Python.
\end{itemize}
Additionally, we explicitly instruct the LLM to report the modifications it makes (\textit{``Include detailed docstrings explaining: ...''}). This step is essential, as it enables us---as shown in Section~\ref{sec:evaluation}---to understand why a particular LLM' code outperforms the original or another model's code.

\section{Experimental evaluation}\label{sec:evaluation}
In this section, we present experiments with the 10 previously mentioned optimization algorithm codes taken from  \texttt{pyCombinatorial}. We describe our setup for utilizing LLMs, the parameter tuning of the stochastic algorithms, the TSP datasets and evaluation metrics employed, and the comparative analysis of the results. We highlight key details from the generation process and conclude with an analysis concerning code complexity.

\subsection{Setup}

\subsubsection{LLMs Environment}

We selected five leading code-generation LLMs: Anthropic Claude-3.5-Sonnet~\cite{anthropicIntroducingClaude}, Google Gemini-exp-1206~\cite{geminiteam2024gemini15unlockingmultimodal}, Meta Llama-3.3-70b~\cite{grattafiori2024llama3herdmodels}, OpenAI GPT-O1~\cite{openai2024gpt4technicalreport}, and DeepSeek-R1~\cite{deepseekai2025deepseekr1incentivizingreasoningcapability}. For simplicity, we will refer to the models as Claude, Gemini, Llama, O1, and R1 throughout the remainder of this paper. Note that these LLMs rank among the top models in the LiveBench benchmark~\cite{white2024livebenchchallengingcontaminationfreellm}, which is immune to both test set contamination and the biases of LLM-based and human crowdsourced evaluations (\textbf{as of February 2025}). Using the OpenRouter API, we executed identical prompts across all models, enabling straightforward model switching for transparent experimentation. This produced 50 new algorithm codes which, combined with the 10 original algorithm codes from the \texttt{pyCombinatorial} framework, gave us 60 Python files ready for evaluation.

\subsubsection{Hardware Environment}\label{subsec:hardware}

All experiments, including parameter tuning, are conducted on a cluster equipped with Intel® Xeon® CPU 5670 processors (12 cores at 2.933 GHz) and 32 GB of RAM.

\subsubsection{Parameter Tuning}
  
While the deterministic heuristics and the branch and bound method are parameter-less, the seven probabilistic approaches (five metaheuristics and two reinforcement learning algorithms) require careful parameter tuning to perform well. Consequently, we tuned all 42 stochastic algorithm codes: the 7 original versions and the 35 new variants generated by the LLMs. To ensure a fair and robust comparison, we employed \texttt{irace}~\cite{irace}, a well-established automatic algorithm configuration tool. The tuning process was executed as parallel jobs on the SLURM cluster described in Section~\ref{subsec:hardware}. For each algorithm execution during the tuning phase, the CPU time limit was set to the number of cities in the instance (in seconds). Table~\ref{tab:tuning} details the parameter ranges considered for tuning each algorithm variant, as well as the final best configurations selected by \texttt{irace}.

\begin{tcolorbox}[colback=bluesky!10,  boxrule=0.2pt, breakable]
Parameter tuning in stochastic algorithms with parameters is essential, as suboptimal configurations can lead to poor performance regardless of the algorithm's inherent quality. For instance, in ACO, we tune key parameters---$m$ (number of ants), $\alpha$, $\beta$, and $\rho$ (pheromone decay)---for all six code variants (five LLM-generated ones plus the original) to ensure they operate under optimal conditions for the TSP problem. This guarantees a fair evaluation, as each code variant is assessed using its best possible configuration.
\end{tcolorbox}
\begin{table*}[t]
    \centering
    \scriptsize
    \caption{Parameter values obtained by tuning with \texttt{irace}. Ranges show minimum/maximum values considered for tuning.}
    \resizebox{\linewidth}{!}{ 
    
    \begin{tabular}{l|lc|ccccccc}
    \toprule
    \textbf{Algorithm} & \textbf{Parameter} & \textbf{Range} & \textbf{Original} & \textbf{Claude-3.5-Sonnet} & \textbf{Gemini-exp-1206} & \textbf{Llama-3.3-70B} & \textbf{GPT-O1} & \textbf{DeepSeek-R1} \\
    \midrule
    \multicolumn{9}{l}{\textcolor[RGB]{70, 130, 180}{\textbf{Metaheuristics}}} \\
    ACO & $m$ (ants) & (2, 20) & 7 & 4 & 2 & 3 & 17 & 20 \\
                           & $\alpha$ (alpha) & (1.0, 2.0) & 1.34 & 1.72 & 1.67 & 1.46 & 1.72 & 1.22 \\
                           & $\beta$ (beta) & (1.0, 2.0) & 1.59 & 1.24 & 1.98 & 1.97 & 1.93 & 1.55 \\
                           & $\rho$ (decay) & (0.01, 0.3) & 0.24 & 0.12 & 0.24 & 0.29 & 0.06 & 0.05 \\
    \midrule
    GA & $N$ (population size) & (5, 100) & 97 & 14 & 97 & 84 & 55 & 58 \\
                         &  $\mu$ (mutation rate) & (0.01, 0.2) & 0.02 & 0.04 & 0.16 & 0.16 & 0.01 & 0.13 \\
                         & $e$ (elite) & (1, 5) & 4 & 2 & 5 & 2 & 3 & 5 \\
    \midrule
    ALNS & $\lambda$ (removal fraction) & (0.05, 0.3) & 0.27 & 0.05 & 0.26 & 0.29 & 0.22 & 0.29 \\
                              & $\rho$ (rho) & (0.01, 0.3) & 0.27 & 0.25 & 0.04 & 0.2 & 0.27 & 0.02 \\
    \midrule
    
    TABU & $T$ (tabu tenure) & (3, 30) & 8 & 12 & 30 & 15 & 10 & 9 \\
    \midrule
    SA & $T_0$ (initial temperature) & (1, 50) & 12 & 49 & 9 & 30 & 35 & 50 \\
                        & $T_f$ (final temperature) & (0.0001, 0.1) & 0.0547 & 0.0464 & 0.074 & 0.056 & 0.0433 & 0.048 \\
                        & $\alpha$ (cooling rate) & (0.8, 0.99) & 0.9895 & 0.8732 & 0.8956 & 0.8131 & 0.9154 & 0.8777 \\
    \midrule
    \multicolumn{9}{l}{\textcolor[RGB]{70, 130, 180}{\textbf{Reinforcement Learning}}} \\
    RL\_QL & $lr$ (learning rate) & (0.01, 0.5) & 0.44 & 0.15 & 0.26 & 0.49 & 0.46 & 0.34 \\
                        & $df$ (decay factor) & (0.8, 0.99) & 0.97 & 0.82 & 0.98 & 0.87 & 0.98 & 0.82 \\
                        & $\epsilon$ (epsilon) & (0.01, 0.3) & 0.09 & 0.28 & 0.03 & 0.24 & 0.21 & 0.13 \\
                        & $E$ (episodes) & (1000, 10000) & 4266 & 1082 & 4906 & 2474 & 1294 & 1989 \\
    \midrule
    SARSA & $lr$ (learning rate) & (0.01, 0.5) & 0.04 & 0.36 & 0.49 & 0.19 & 0.41 & 0.29 \\
                          & $df$ (decay factor) & (0.8, 0.99) & 0.86 & 0.91 & 0.80 & 0.88 & 0.83 & 0.87 \\
                          & $\epsilon$ (epsilon) & (0.01, 0.3) & 0.23 & 0.18 & 0.16 & 0.08 & 0.12 & 0.16 \\
                          & $E$ (episodes) & (100, 5000) & 105 & 156 & 1850 & 137 & 124 & 1711 \\
    \bottomrule
\end{tabular}
\label{tab:tuning}
    }
\end{table*}

\subsection{Benchmark Datasets and Evaluation Metrics}
To evaluate all algorithm codes, we use problem instances from the well-known TSPLib library~\cite{journals/informs/Reinelt91}. We select 10 instances from the available ones, ranging from a small instance with 99 cities to a large one with 1084 cities.\footnote{TSPLib names of the selected TSP instances: \textsc{rat99}, \textsc{bier127}, \textsc{d198}, \textsc{a280}, \textsc{f417}, \textsc{ali535}, \textsc{gr666}, \textsc{u724}, \textsc{pr1002}, and \textsc{vm1084}.} This selection ensures a comparison across a diverse set of problem instance sizes.

As an evaluation metric, we used the objective function value of the best-found solution in all cases except for the Branch and Bound (BB) codes. This is because BB is an exact algorithm that, if given enough computation time, will always find an optimal solution. Therefore, we use runtime as the evaluation metric in the case of BB. Moreover, as the runtime of BB for the 10 selected problem instances is very high, we instead generate 10 random TSP instances with 10 to 15 cities for the evaluation of the BB codes. Interestingly, as we will see in the comparative analysis subsection, some LLM-generated versions of BB incorporate heuristic mechanisms during algorithm initialization, leading to significant improvements in runtime performance.

\subsection{Experimental Design}\label{subsec:exp-design}

\noindent The experiments were designed as follows:
\begin{itemize}
    \item \textbf{Stochastic Algorithms}. Each of the metaheuristics and reinforcement learning codes is applied 30 times independently to each of the 10 problem instances. The output of each run is the best solution found. Performing 30 independent algorithm executions for each problem instance is a common practice in the optimization community to obtain a reliable estimate of the algorithm's performance. Moreover, the CPU time limit for each algorithm execution is set to the number of cities (in seconds) of the tackled problem instance. For example, the run-time limit for \textsc{rat99} is 99 seconds. This method, which aligns execution time with the instance size, is a common practice for comparing algorithms that solve the TSP.    
    \item \textbf{Deterministic Heuristics}. Christofides and Convex Hull, since they are deterministic heuristics, always yield the same result for a given problem instance. Therefore, all corresponding codes are executed exactly once per instance.
    \item \textbf{Branch and Bound}. As previously mentioned, since Branch and Bound is an exact algorithm, the focus is on runtime rather than solution quality. All Branch and Bound codes are applied 30 times to each of the small problem instances specifically generated for the evaluation of the Branch and Bound codes.
\end{itemize}

\subsection{Comparative Analysis with Original Algorithm Codes}

The comparative analysis between the LLM-updated algorithm codes and the original algorithm codes (referred to as `original' from now on) is studied in the following in a separate way depending on the type of optimization algorithm.

\subsubsection{Metaheuristics}

\begin{figure*}[t]
    \centering
    \includegraphics[width=\linewidth]{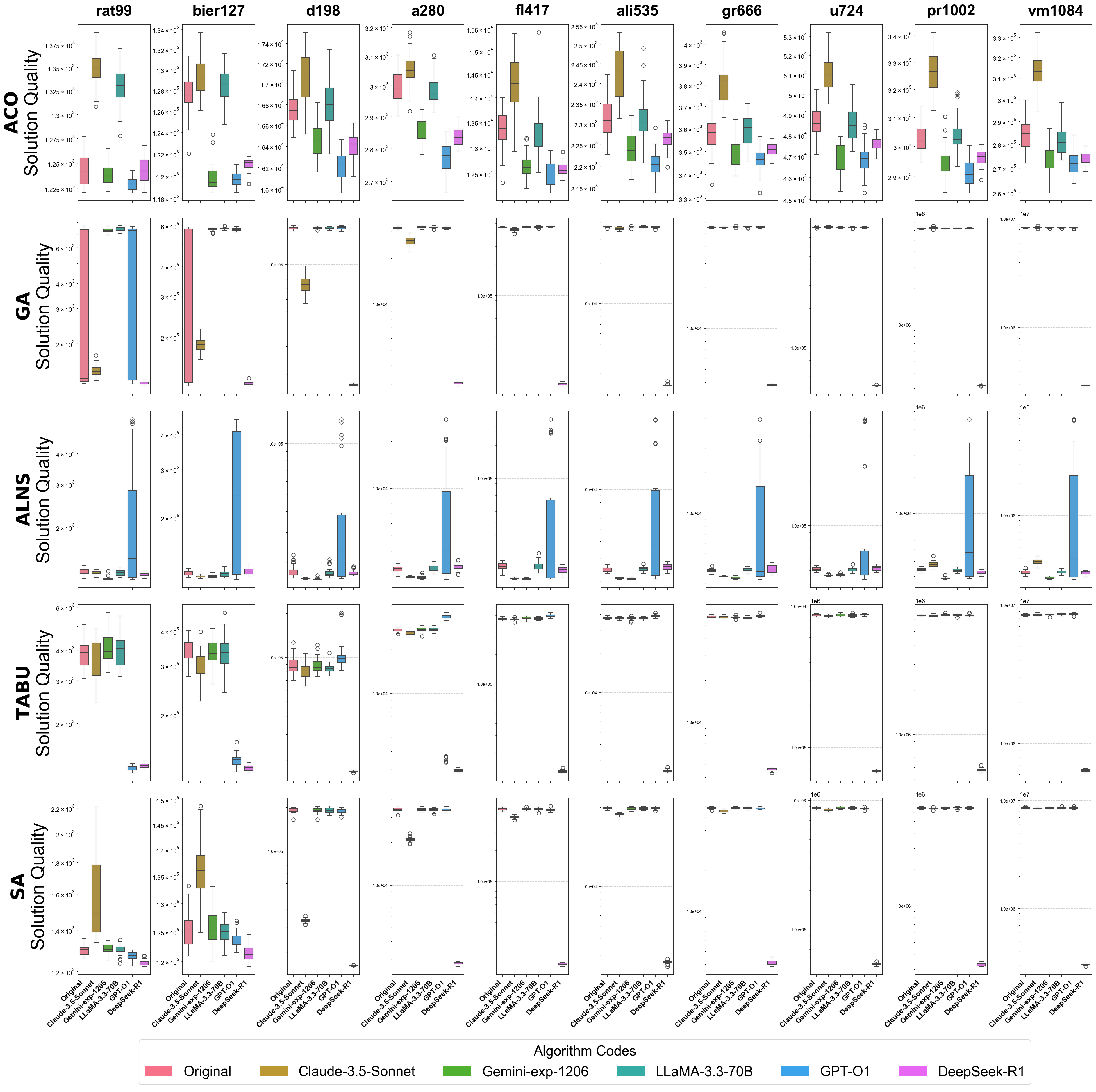}
    \caption{Comparison of the metaheuristic codes generated by the five LLMs with the original codes. Remember that the TSP is a minimization problem, that is, the lower the values, the better. The y-axes are shown in a logarithmic scale.}
    \label{fig:results-metaheuristic}
\end{figure*}

The results of all metaheuristic codes are shown by means of boxplots in Figure~\ref{fig:results-metaheuristic}. Note that the y-axes are shown in a logarithmic scale.
\begin{enumerate}
    \item \textbf{ACO}: Apart from the first problem instance (\textsc{rat99}) where the LLM-generated codes of Gemini, O1, and R1 perform similarly to the original code, in all other problem instances the LLM-generated codes of the mentioned three LLMs outperform the original code with statistical significance. It also appears that the LLM-generated codes of O1 and R1 are somewhat more robust than the original code, which can be seen in smaller boxes. In contrast, the code generated by Claude, apart from the first problem instance, performs always worse than the original code. Finally, the Llama-generated code generally performs similarly to the original code, with the exception of the first problem instance.
    
    
    \item \textbf{GA}: The original code exhibits very low robustness for the first two problem instances, which can be seen by the large boxes. Generally, most codes (except for R1) are quickly trapped in local optima which they cannot escape. The R1-generated code clearly outperforms all others, including the original.

    \item \textbf{ALNS}: Generally, the best-performing codes are those by Claude and Gemini, with a slight advantage for Gemini in the last two instances. Another noteworthy aspect is the low robustness of the O1-generated code in this case.

    \item \textbf{TABU}: Like in the case of GA, also the TABU code generated by R1 significantly outperforms the remaining codes. Only the O1-generated code can compete for the smallest two problem instances. This suggests that R1 excels at generating efficient optimization algorithms.

    \item \textbf{SA}: The Claude-generated codes clearly show the weakest performance here. In contrast, the R1-generated code again outperforms the remaining ones. 
\end{enumerate}

\subsubsection{Reinforcement Learning (RL)}
\begin{figure*}[t]
\centering
    \includegraphics[width=\linewidth]{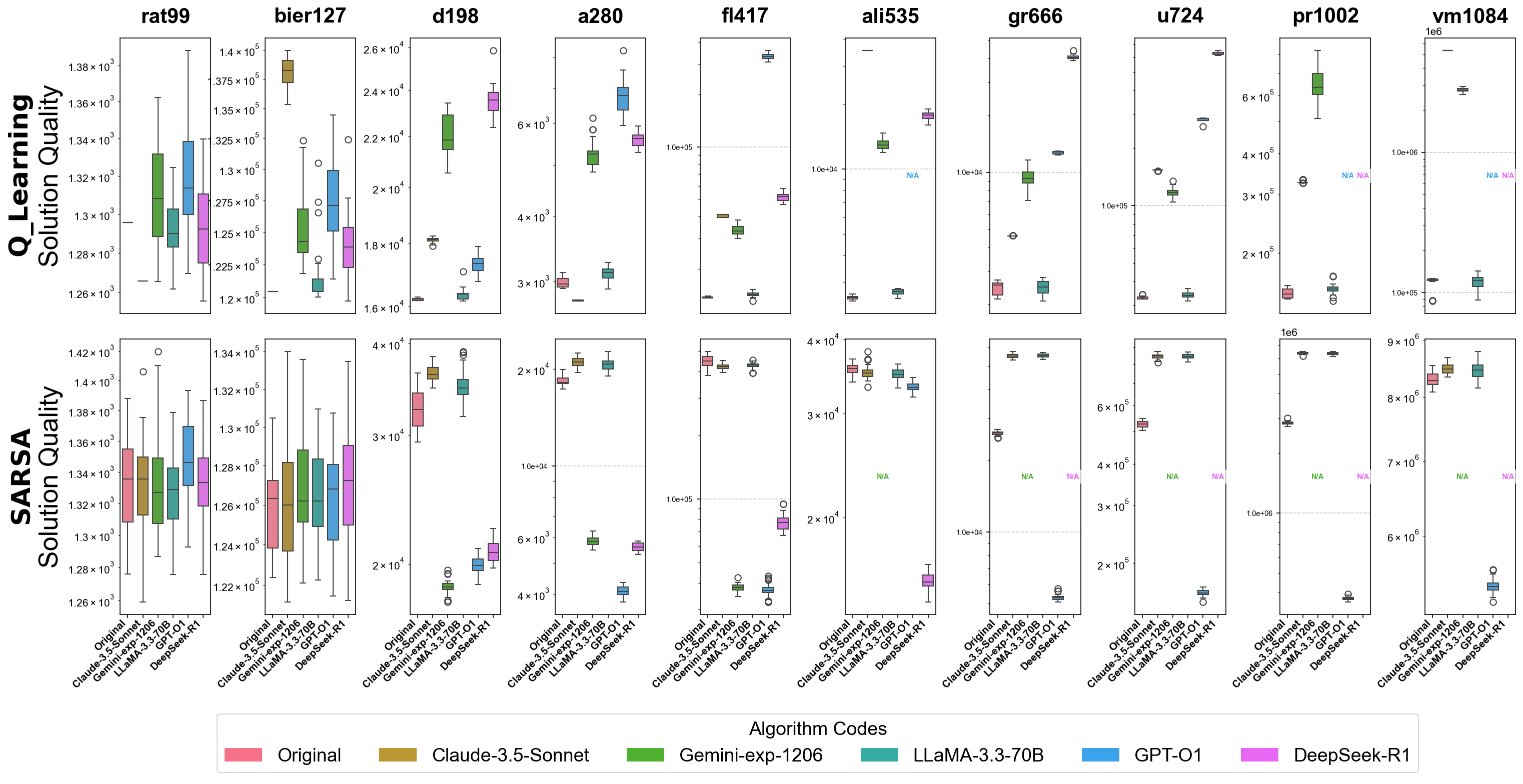}
    \caption{Comparison of the reinforcement learning (RL) codes generated by the five LLMs with the original codes. Remember that the TSP is a minimization problem, that is, the lower the values, the better. The y-axes is shown in logarithmic scale.}
    \label{fig:results-rl}
\end{figure*}

In Figure~\ref{fig:results-rl}, the RL codes generated by the LLMs are compared with the original RL codes. The displayed results differ from the metaheuristics case presented before in two key aspects. First, some LLM-generated codes fail to produce a result for all problem instances within the time limit. These cases are marked as \textsc{N/A}. Second, especially in the case of Q\_Learning the original code performs more competitively. We analyze these aspects below:

\begin{enumerate}
\setcounter{enumi}{5}
    \item \textbf{Q\_learning}: The original code is actually the best-performing one in this case. The Llama-generated code is the only one that achieves a nearly comparable performance---an unexpected result given Llama's poor performance in the case of the metaheuristics. In addition, the Claude-generated code outperforms the original one on the \textsc{rat99} and \textsc{a280} instances.
    
    \item \textbf{SARSA}: The general picture here aligns more with that observed in the case of the metaheuristics. The original code struggles (in comparison to some LLM-generated codes) as instance sizes grow larger. The R1-generated code fails to produce results for the largest three instances. The only code maintaining a stable and strong performance across all 10 instances is the O1-generated one.
\end{enumerate}

\subsubsection{Deterministic/Heuristic}  

\begin{figure*}[t]
    \centering
    \begin{minipage}{0.48\linewidth}
        \centering
        \includegraphics[width=1\linewidth]{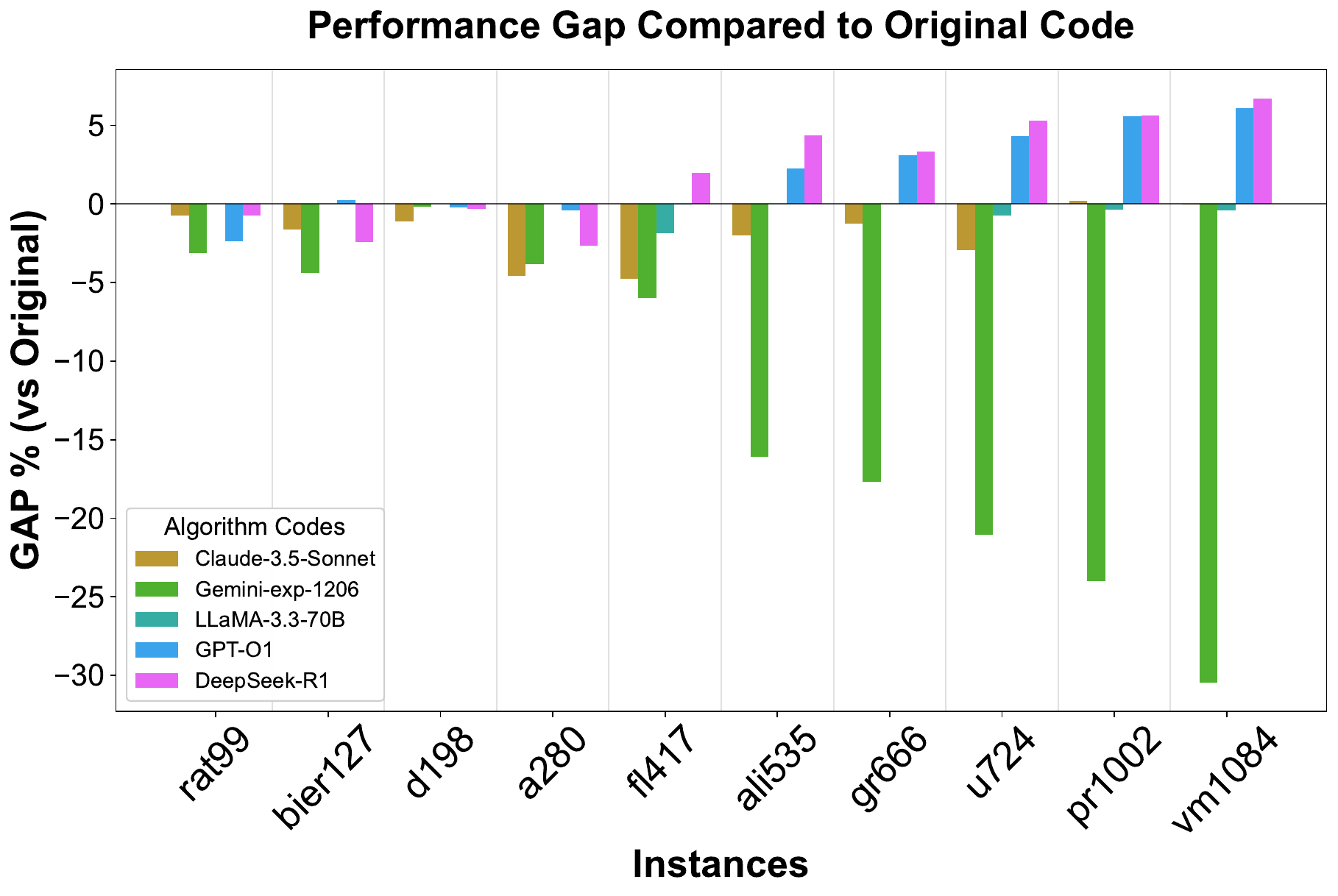}
        \parbox{\linewidth}{\centering (a) Christofides}
    \end{minipage}
    \hfill
    \begin{minipage}{0.48\linewidth}
        \centering
        \includegraphics[width=1\linewidth]{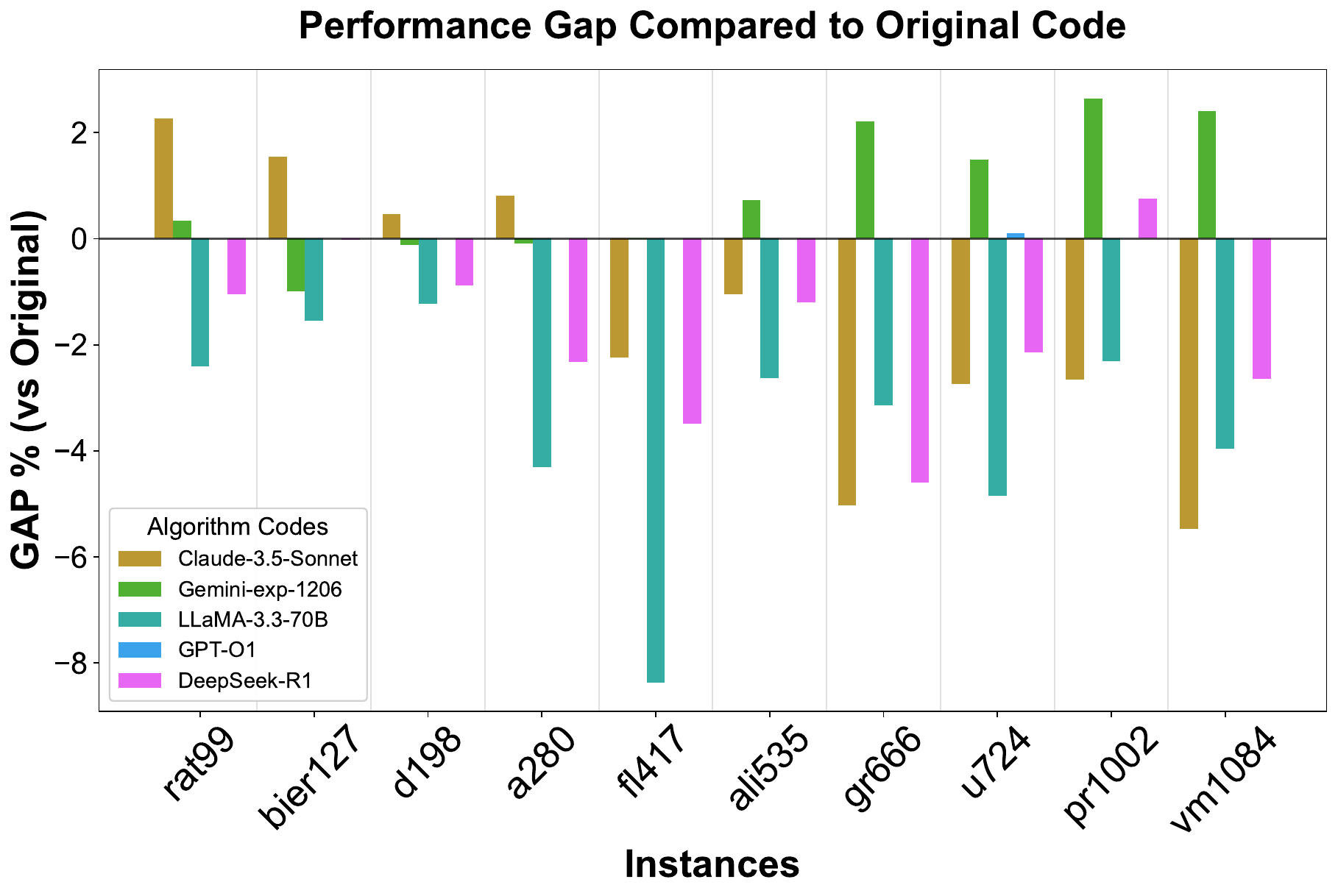}
        \parbox{\linewidth}{\centering (b) Convex Hull}
    \end{minipage}
    
    \caption{Comparison of the deterministic heuristic codes generated by the five LLMs with the original codes. The bar plots show the performance gaps (in percent) relative to the original codes. Note that a positive value indicates that the LLM-generated code produces a better solution.}
    \label{fig:results-heuristics}
\end{figure*}

Remember that, as the chosen heuristics are deterministic, there is no need to analyze the distribution of their results over multiple runs. Therefore, in Figure~\ref{fig:results-heuristics}, we simply compare the GAP of the results produced by the LLM-generated codes (in percent) relative to the results of the original codes. A positive value indicates that the respective LLM-generated code outperforms the original, while a negative value suggests the opposite.

\begin{enumerate} \setcounter{enumi}{7}
\item \textbf{Christofides} (see Figure~\ref{fig:results-heuristics}~(a)): While the Llama-generated code produces very similar results to the original code over the whole instance range, the Gemini-generated code is (apart from instance \textsc{d198}) always inferior. Moreover, its relative performance decreases as instance size grows. Concerning O1 and R1, it can be stated that the performance of their codes is slightly inferior to the one of the original code for rather small problem instances. However, with growing instance size, they clearly outperform the original code.


\item \textbf{Convex Hull} (see Figure~\ref{fig:results-heuristics}~(b)): In contrast to Christofides, the Convex Hull codes generated by O1 and R1 perform rather poorly. In fact, the best code for Convex Hull is the one generated by Gemini. This code has slight disadvantages for smaller instances but increasingly outperforms the original code with growing instance size. The code generated by Claude shows the opposite pattern. While it outperforms the original code for smaller problem instances, its performance strongly decreases with growing instance size.

\end{enumerate}

\subsubsection{Exact Approach}  

As already mentioned in Section~\ref{subsec:exp-design} (Experimental Design), in the context of the exact BB method, the comparison is based on computation time.

\begin{figure}[H]
    \centering
    \includegraphics[width=0.5\linewidth]{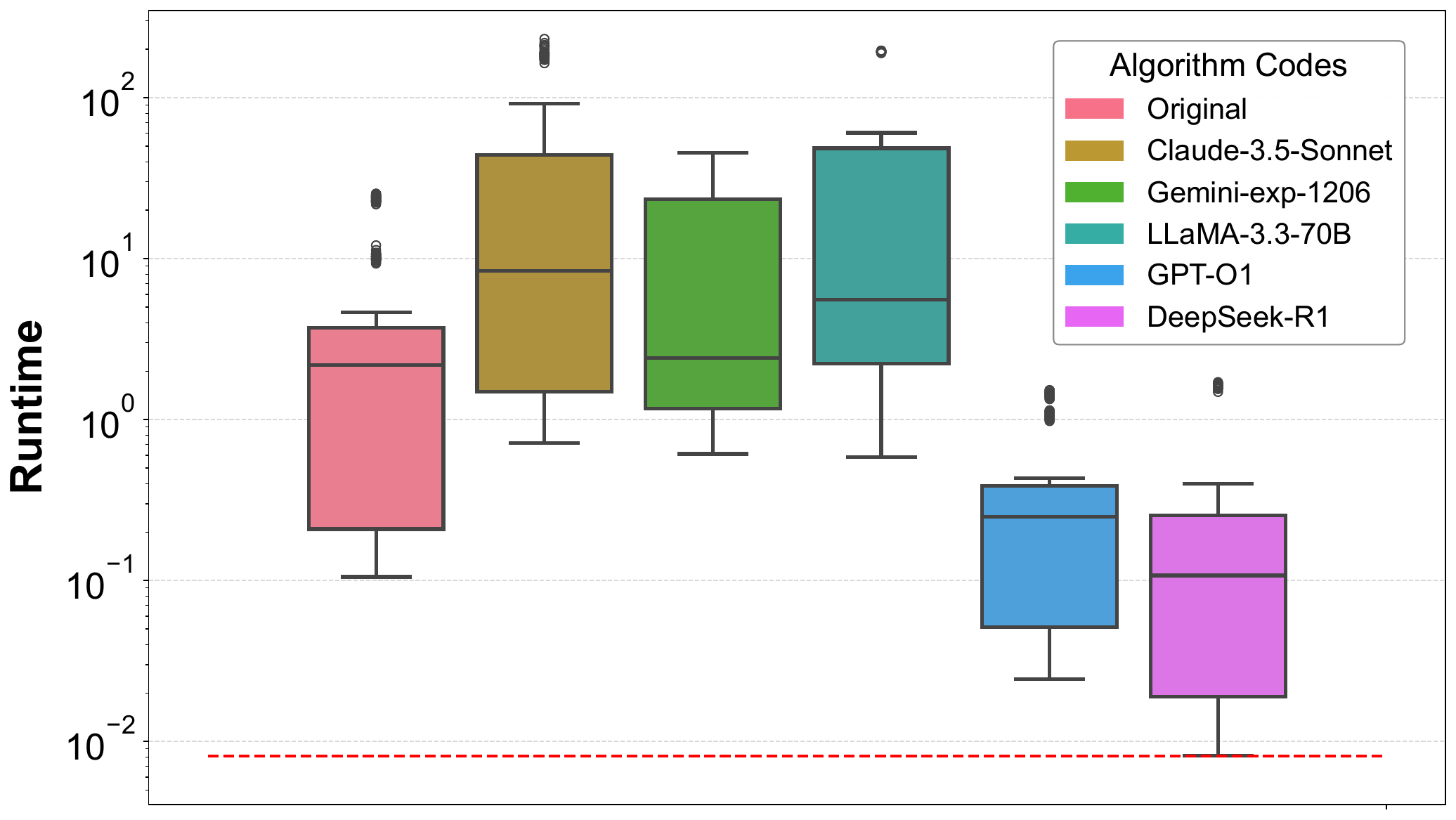}
    \caption{Comparison of the BB codes generated by the five chosen LLMs with the original BB code (in terms of computation time). Each code was applied 30 times, and the y-axis is shown in a logarithmic scale.}
    \label{fig:results-bb}
\end{figure}

\begin{enumerate}
\setcounter{enumi}{9}
    \item \textbf{Branch and Bound}: Figure~\ref{fig:results-bb} shows that the codes generated by O1 and R1 outperform both the original code. Notably, R1---an open-weight LLM---achieves the best performance, surpassing all proprietary models. In contrast to O1 and R1, the other LLM-generated codes perform worse than the original code. 
\end{enumerate}  

\begin{tcolorbox}[colback=bluesky!10,  boxrule=0.2pt, breakable]

\textbf{Summary:} A notable conclusion is that LLMs can produce improved versions of baseline algorithms, resulting in performance improvements without necessitating specialized expertise in each algorithm. In the following subsection, we showcase examples of code improvements achieved, for example, by integrating more sophisticated algorithmic components.
\end{tcolorbox}

\subsection{Key Insights in Code Generation}  

Next, we explore why certain LLM-generated (or LLM-updated) codes outperform the original ones. Our focus was to understand if this was due to optimized data structures, for example, or due to adding different algorithmic components. In particular, we analyze four cases to address these questions on the basis of the LLM-generated codes.

\subsubsection{Case 1: GA (R1-generated code)}\label{case1}  

The R1-generated version of GA features the following improvements, as stated by the model itself by means of a docstring in the code, as requested in the prompt.
    
\begin{lstlisting}[language=Python,
    backgroundcolor=\color{backcolour},   
    commentstyle=\color{codegreen},
    keywordstyle=\color{codepurple},
    numberstyle=\tiny\color{codegray},
    stringstyle=\color{codepurple},
    basicstyle=\ttfamily\footnotesize,
    breakatwhitespace=false,         
    breaklines=true,                 
    keepspaces=true,              
    showspaces=false,       
    showstringspaces=false,
    showtabs=false,
    tabsize=2]
Improvements:
1. Hybrid initialization with nearest neighbor heuristic
2. Rank-based fitness + tournament selection
3. Adaptive operator selection (OX, ER, BCR)
4. Memetic local search with stochastic 2-opt
5. Diversity preservation mechanisms
\end{lstlisting}
    
In particular, R1-generated GA is the only LLM-generated code that introduces a modification to the population initialization by incorporating the nearest neighbor heuristic. In contrast, both the original code and all other LLM-generated variants use the following initialization function:

\begin{lstlisting}[language=Python,
    backgroundcolor=\color{backcolour},   
    commentstyle=\color{codegreen},
    keywordstyle=\color{codepurple},
    numberstyle=\tiny\color{codegray},
    stringstyle=\color{codepurple},
    basicstyle=\ttfamily\scriptsize,  % Cambiado a scriptsize o \tiny si quieres aún más pequeño
    breakatwhitespace=false,         
    breaklines=true,                 
    keepspaces=true,                  
    showspaces=false,       
    showstringspaces=false,
    showtabs=false,
    tabsize=2,]
# Function: Initial Population
def initial_population(population_size, distance_matrix):
    population = []
    for i in range(0, population_size):
        seed = seed_function(distance_matrix)
        population.append(seed)
    return population

\end{lstlisting}

Instead, R1-generated GA features the following initialization that leads to an improved performance.\footnote{Note that, in all Python code snippets shown in this paper, `...' indicates omitted parts that are not relevant.}

\begin{lstlisting}[language=Python,
    backgroundcolor=\color{backcolour},   
    commentstyle=\color{codegreen},
    keywordstyle=\color{codepurple},
    stringstyle=\color{codepurple},
    basicstyle=\ttfamily\footnotesize,
    basicstyle=\ttfamily\scriptsize,
    breakatwhitespace=false,         
    breaklines=true,                 
    keepspaces=true,                 
    numbers=left,                   
    showspaces=false,       
    showstringspaces=false,
    showtabs=false,
    tabsize=1,
    numbersep=2pt]
def initial_population(population_size, distance_matrix):
    """Initialize population with mix of random and heuristic solutions. Combines diversity (random) with quality (NN) for better exploration. Implements hybrid population initialization from modern metaheuristics."""
    population = []
    if population_size >= 5:  # Include 20% NN seeds
        for _ in range(max(1, population_size//5)):
            population.append(nearest_neighbor_seed(distance_matrix))
    
    ...
    return population
def nearest_neighbor_seed(distance_matrix):
    """Generate initial solution using Nearest Neighbor heuristic. Provides high-quality initial seeds to accelerate convergence. Based on constructive heuristic methods commonly used in TSP."""
    ...
\end{lstlisting}
In particular, the GA is initialized with 20\% nearest neighbor solutions for population sizes of at least five individuals. This well-known TSP heuristic significantly speeds up convergence. In this way, R1 shows its ability to `dig' into its knowledge base to choose an alternative population initialization method and implement it effectively.


\subsubsection{Case 2: SA (R1-generated code)}\label{case2}  
Also in the case of SA, R1 identifies and utilizes two well-known mechanisms recognized for their efficiency in solving the TSP: (1) the Lundy-Mees adaptive cooling schedule for improved temperature control, introduced years after the original SA~\cite{Lundy1986}, and (2) the nearest neighbor heuristic for TSP. In the latter case, R1 integrates the nearest neighbor heuristic for the TSP in a way similar to what it did in the case of the GA, demonstrating a consistent pattern in leveraging effective initialization strategies.

\subsubsection{Case 3: SARSA (O1-generated code)}  
The O1-generated SARSA code achieved the best results among the competitors. This is due to being the only code to make use of \textit{Boltzmann Exploration} (see code below). Unfortunately, LLMs do not have the capacity to identify the exact source (book, scientific article, etc) from which the information about Boltzmann Exploration was extracted. However, after reviewing the code, it is likely that it was sourced from a 2017 paper (see~\cite{asadi2017alternativesoftmaxoperatorreinforcement}), which suggests a Boltzmann operator for SARSA applied to the TSP.

In fact, the code below shows that, unlike the original code, O1 not only applies a random operator to select the next unvisited city but also assigns a probability---derived from the \texttt{q\_table} data structure---to this choice (line 8), making the selection more dynamic. Moreover, it avoids unnecessary abstractions (e.g., extra data structures) that could slow down the Python code.

\begin{lstlisting}[language=Python,
    backgroundcolor=\color{backcolour},   
    commentstyle=\color{codegreen},
    keywordstyle=\color{codepurple},
    stringstyle=\color{codepurple},
    basicstyle=\ttfamily\footnotesize,
    basicstyle=\ttfamily\scriptsize,
    breakatwhitespace=false,         
    breaklines=true,                 
    keepspaces=true,                 
    numbers=left,                   
    showspaces=false,       
    showstringspaces=false,
    showtabs=false,
    tabsize=1,
    numbersep=2pt]
...
while len(visited) < num_cities:
    unvisited = [city for city in range(num_cities) if city not in visited]
    # Boltzmann exploration
    q_values = q_table[current_city, unvisited]
    exp_q = np.exp(q_values / temperature)
    probabilities = exp_q / np.sum(exp_q)
    next_city = np.random.choice(unvisited, p=probabilities)

    reward = -distance_matrix_normalized[current_city, next_city]
    visited.add(next_city)
    route.append(next_city)
    ...
...
\end{lstlisting}

\subsubsection{Case 4: BB (R1-generated code)}
When studying why the BB code of R1 was faster than the original code, first we noticed that, like in cases~\ref{case1} and~\ref{case2}, R1 made use of the nearest neighbor heuristic for initialization. Moreover, R1 modified the \texttt{explore\_path} function of BB by dynamically sorting the next candidates by edge weight to prioritize the cheapest/nearest extensions first. Both updates are not trivial. R1 notes the following in the code comments: \textit{``Enhancements reduce unnecessary branching and accelerate convergence through early solution bias.''}~\footnote{This can be seen in line~48 of file \texttt{bb\_deepseek\_r1.py} of our online repository URL.}

In addition, the code snippet below is not present in the original code. O1 introduces \texttt{current\_node} and \texttt{candidates} efficiently, using slicing (line~4) and sorting with a lambda function (line~5) to enhance path exploration in BB. This new array-based data structure is both efficient and implemented in a Pythonic style to improve performance.

\begin{lstlisting}[language=Python,
    backgroundcolor=\color{backcolour},   
    commentstyle=\color{codegreen},
    keywordstyle=\color{codepurple},
    stringstyle=\color{codepurple},
    basicstyle=\ttfamily\footnotesize,
    basicstyle=\ttfamily\scriptsize,
    breakatwhitespace=false,         
    breaklines=true,                 
    keepspaces=true,                 
    numbers=left,                   
    showspaces=false,       
    showstringspaces=false,
    showtabs=false,
    tabsize=1,
    numbersep=2pt]
def explore_path(route, distance, distance_matrix, bound, weight, level, path, visited, min1_list, min2_list):
    ...
    current_node = path[level - 1]
    candidates = [i for i in range(distance_matrix.shape[0]) if distance_matrix[current_node, i] > 0 and not visited[i]]
    candidates = sorted(candidates, key=lambda x: distance_matrix[current_node, x])
    ...
\end{lstlisting}

\subsection{Code complexity}

In the previous subsection, we analyzed the LLM-generated codes in terms of their performance. But do these codes also offer better readability and reduced complexity in comparison to the original codes? To address this question, we evaluate their \textit{cyclomatic complexity}---a metric that quantifies the number of independent paths through a program's source code. Through empirical research, Chen~\cite{chen2019empiricalinvestigationcorrelationcode} demonstrated that a high cyclomatic complexity correlates with an increased bug prevalence. For our measurements, we employ the \textsc{Radon} library for Python.\footnote{\url{https://pypi.org/project/radon/}.}

As shown in Table~\ref{tab:avg_complexity}, the Claude-generated codes have the lowest average cyclomatic complexity (5.60 points), which improves code readability but, as shown before, comes at the cost of performance. The other models' codes and the original code have complexity scores between 6.84 (O1) and 7.51 (R1), which is still considered low and well-structured according to standard software engineering metrics. The values in the Risk Category column are taken from the documentation of the \textsc{Radon} library.\footnote{\href{https://radon.readthedocs.io/en/latest/commandline.html\#the-cc-command}{https://radon.readthedocs.io/en/latest/commandline.html\#the-cc-command}}

\begin{figure}[H]
    \centering
    \includegraphics[width=0.7\linewidth]{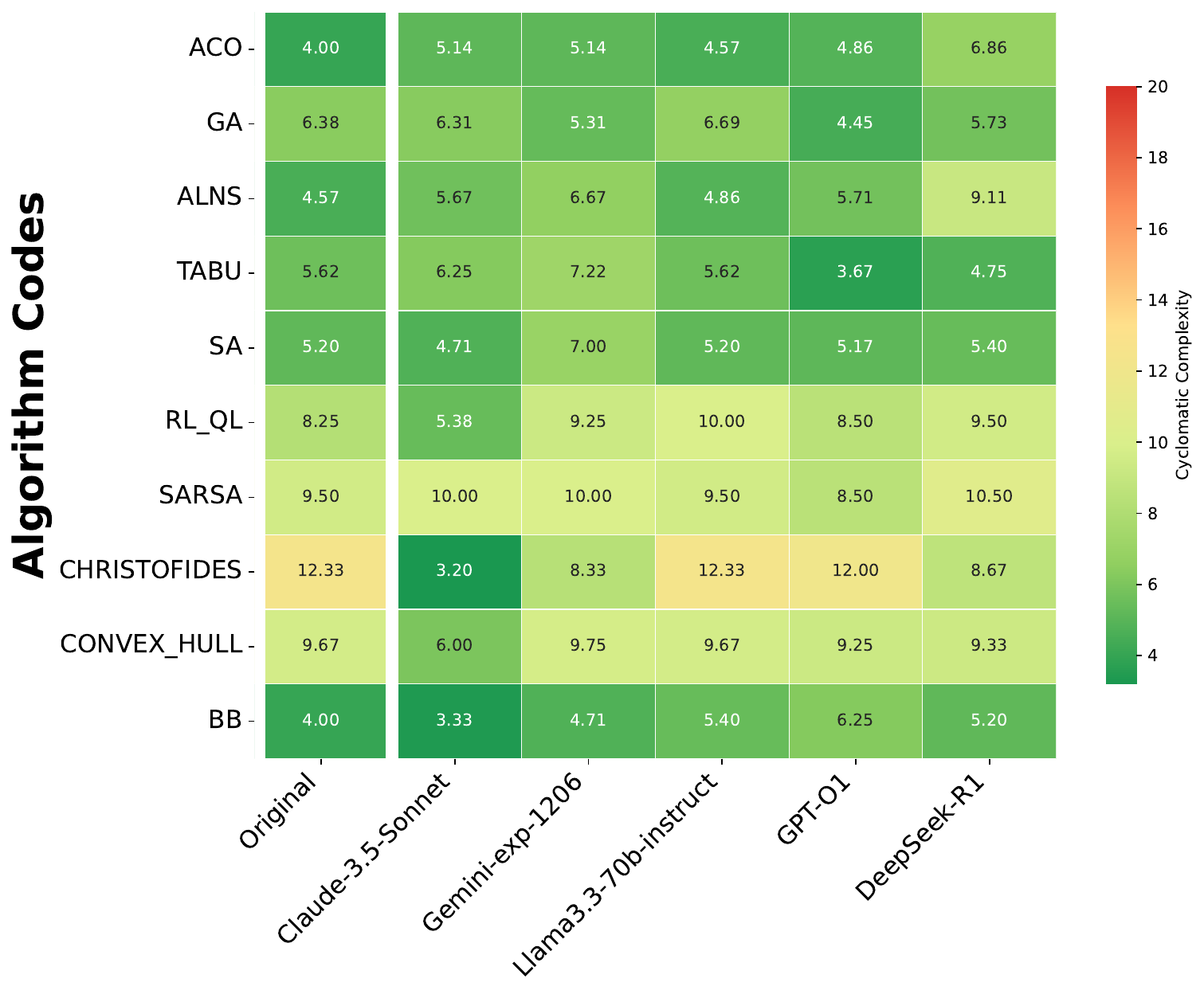}
    \caption{Cyclomatic Complexity}
    \label{fig:results-codecomplexity}
\end{figure}

Finally, Figure~\ref{fig:results-codecomplexity} reveals that there are cases---such as the R1-generates codes in the case of GA and Christofides, or the O1-generated code for SARSA---in which the LLM-generated codes not only outperform the original codes, but also decrease the cyclomatic complexity.

\begin{table}[h]
\centering
\caption{Average Cyclomatic Complexity of the codes}
\label{tab:avg_complexity}
\renewcommand{\arraystretch}{1.5}  
\begin{tabular}{c l c p{0.5\columnwidth}}
\hline
\textbf{} & \textbf{Algorithm Codes} & \textbf{Average Complexity} & \textbf{Risk Category} \\
\hline
           & Original               & 6.95 & B (Low - Well structured) \\  
\hline
\multirow{5}{*}{\rotatebox[origin=c]{90}{LLMs}}  
           & Claude-3.5-Sonnet      & 5.60 & A (Low - Simple) \\  
           & Gemini-exp-1206        & 7.34 & B (Low - Well structured) \\  
           & Llama-3.3-70b          & 7.38 & B (Low - Well structured) \\  
           & GPT-O1                 & 6.84 & B (Low - Well structured) \\  
           & DeepSeek-R1            & 7.51 & B (Low - Well structured) \\  
\hline
\end{tabular}
\end{table}

\begin{tcolorbox}[colback=bluesky!10,  boxrule=0.2pt, breakable]
\textbf{Summary:} Based on all evaluations presented in this paper, we can state that among the five LLMs tested, R1 generally produces the best results, followed by O1. Gemini performed well in certain cases, such as ACO and ALNS; however, it underperformed in others, such as, for example, Christofides. Among all tested models, Claude showed the lowest performance.

In summary, \textit{LLM-enhanced code versions clearly outperformed the original implementations in nine out of ten cases/algorithms}. Only for Q\_Learning none of the models was able to improve the original code. In this case, Llama matched the performance of the original code.

\end{tcolorbox}

\section{Discussion}\label{sec:disc}

  Recent research has convincingly demonstrated that LLMs can be powerful tools for creating and enhancing complex optimization algorithms~\cite{sartori2025improvingexistingoptimizationalgorithms, Novikov2025-vg, 10752628, ye2024reevolargelanguagemodels, Romera-Paredes2024, liu2024evolutionheuristicsefficientautomatic}. To situate our current work within this rapidly evolving field and clearly delineate its unique contributions, we provide a comprehensive comparative analysis of these state-of-the-art approaches in Table~\ref{tab:expanded_comparative_analysis}. Building upon our own initial proof-of-concept~\cite{sartori2025improvingexistingoptimizationalgorithms}, this work significantly expands the investigation by conducting a large-scale, systematic study across 10 classical algorithms from four diverse families. Our findings confirm that the principle of LLM-driven improvement is not an isolated phenomenon but a broadly applicable technique for the Travelling Salesman Problem. However, our analysis also reveals several critical limitations and methodological considerations that warrant a deeper discussion.

\subsection{Limitations and Methodological Considerations}

While our results are promising, a nuanced understanding of the challenges is crucial for the practical application and future development of this approach.

\begin{itemize}
    \item \textbf{The ``Black Box'' Nature and Its Risks.} A primary challenge, highlighted in~\cite{khalifa2024sourceawaretrainingenablesknowledge}, is that LLMs cannot precisely trace the sources of their suggestions. This opacity makes it difficult to pinpoint which specific modifications led to performance gains. More importantly, it introduces the risk of the LLM generating ``hallucinated'' algorithmic components or incorrectly combining concepts from different sources, potentially leading to subtle logical flaws that are not immediately apparent through basic testing. Exploring the question of \textit{``where specifically does the generated code come from?''} is therefore not just a fascinating research avenue but a critical step towards building more reliable systems.

    \item \textbf{The Practical Costs of Iterative Refinement.} Our ``simple prompt'' strategy successfully lowers the barrier to entry, but the overall process is not without cost. As shown in Table~\ref{tab:generation_algorithms}, several algorithms required multiple manual correction rounds to become functional. This iterative cycle of generating, testing, and providing feedback demands significant user time and effort. This reveals a key trade-off: the ease of initial prompting versus the manual cost of validation. For this methodology to be truly effective in practice, the efficiency of this human-in-the-loop refinement process must be considered.
    \item \textbf{Performance Variability and Failure Analysis.} 
    Our results reveal significant performance variability among LLMs and highlight specific failure cases, as promised in our introduction. For instance, the general failure of all tested LLMs to improve upon the Q-Learning baseline (Figure~\ref{fig:results-rl}) suggests that current models may struggle with algorithms whose performance is highly sensitive to specific state-space representations or reward structures. Similarly, the weaker performance of some models on complex algorithms like ALNS may point to a scarcity of high-quality, specialized training data for such intricate heuristics. Understanding these failure modes is essential for defining the boundaries of where LLM-based improvement is most effective. 
    Conversely, it is particularly notable that improved performance does not always correlate with increased code complexity. In cases such as the GA code generated by R1, our analysis shows that the LLM simultaneously enhanced the solution quality while decreasing the cyclomatic complexity (Figure~\ref{fig:results-codecomplexity}), suggesting that these models can also act as effective code refactorizers---a significant benefit beyond pure algorithmic enhancement.

    \item \textbf{Positioning within the State-of-the-Art.} 
    As detailed in Table~\ref{tab:expanded_comparative_analysis}, our work is methodologically distinct from other recent approaches. While prominent frameworks like LLaMEA/ReEvo, AlphaEvolve, and FunSearch employ complex, automated systems---such as evolutionary loops or program search trees---to generate or discover \textit{new} algorithmic components, our approach uses a simple, interactive prompting strategy. This reflects a fundamental difference in objective and user role: expert-centric frameworks focus on automated algorithm discovery or generation, requiring users to design a sophisticated search process. In contrast, our practitioner-centric method centers on the collaborative enhancement of \textit{complete, existing codebases}. Furthermore, while the work~\cite{sartori2025improvingexistingoptimizationalgorithms} established the feasibility of this concept in a single case study, the present research validates its effectiveness systematically across a broad and diverse range of algorithms.
\end{itemize}

\subsection{Directions for Future Research}

The limitations identified above naturally lead to several exciting directions for future research.

\begin{itemize}
    \item \textbf{Broadening the Problem Scope.} A crucial next step, addressing a limitation of our current study, is to validate these findings beyond the TSP. Applying this methodology to combinatorial optimization problems with different structures, such as the Vehicle Routing Problem (VRP) or the Knapsack Problem, is essential to determine the true generality of this approach.

    \item \textbf{Automating the Refinement Loop.} To mitigate the manual effort of validation and correction, future work should focus on automating the feedback cycle. This could involve developing systems where an LLM agent can not only generate code but also autonomously execute it against a benchmark suite, analyze the results (including errors and performance metrics), and iteratively refine its own suggestions.

    \item \textbf{Specialized Models and Benchmarks.} As LLMs evolve, it will be necessary to continuously test their capabilities in a structured way. The creation of specialized benchmarks with baseline implementations for multiple optimization problems would be invaluable. Furthermore, developing fine-tuned LLMs specialized in optimization could overcome the data scarcity issues observed for complex or less-common algorithms, potentially leading to even more significant and reliable improvements. This aligns with trends towards ``reasoning models'' that prioritize response quality over speed~\cite{snell2024scalingllmtesttimecompute}.
\end{itemize}

\section{Conclusion}\label{sec:conc}

Our systematic benchmark demonstrates that Large Language Models (LLMs) can be powerful collaborators in enhancing classical optimization algorithms. We have shown that a simple, reproducible prompting strategy is sufficient to improve 10 baseline algorithms for the Travelling Salesman Problem, often yielding higher-quality solutions, faster computation times, and reduced code complexity. Crucially, these improvements were achieved holistically on complete codebases through the incorporation of modern heuristics and data structures. This entire methodology is fully reproducible via the chatbot on our project website (\url{https://camilochs.github.io/comb-opt-for-all/}), providing a practical pathway for practitioners to upgrade complex algorithms without deep theoretical expertise.

Building on this foundation, future work will address the limitations identified in our discussion. To validate the generality of our findings, these methods will be extended to other combinatorial optimization problems with diverse structures. Furthermore, to overcome the practical costs of manual refinement, we plan to explore the integration of LLM-based agents to create a fully automated and continuous improvement cycle for existing algorithms.

\section*{Acknowledgements}
The research presented in this paper was supported by grant PID2022-136787NB-I00 funded by MCIN/AEI/10.13039/501100011033. Moreover, thanks to FreePik, from where we extracted the icons used in Figure~\ref{fig:example}. Finally, thanks to Valdecy Pereira, creator of \texttt{pyCombinatorial}, for kindly answering our inquiries.

\bibliographystyle{plainnat}  
\bibliography{main}  

\end{document}